\documentclass[letterpaper, 10 pt, conference]{ieeeconf}  

\IEEEoverridecommandlockouts                              

\usepackage{graphics} 
\usepackage{epsfig} 
\usepackage{mathptmx} 
\usepackage{times} 
\usepackage{amsmath} 
\usepackage{amssymb}  
\usepackage{color}
\usepackage{amsmath}
\usepackage{graphicx, subfigure}
\usepackage{lipsum}
\usepackage{float}
\usepackage{caption}

\title{\LARGE \bf
SLAM based Quasi Dense Reconstruction For Minimally Invasive Surgery Scenes
}

\author{Nader Mahmoud$^{1,2}$, Alexandre Hostettler$^{1}$, Toby Collins$^{1}$, Luc Soler$^{1,3}$, Christophe Doignon$^{2}$, J.M.M. Montiel$^{4}$
\thanks{$^{1}$
IRCAD (Institut de Recherche contre les Cancers de l'Appareil Digestif), France.
   {\tt\small  nader-mahmoud.ali@etu.unistra.fr}}%
\thanks{$^{2}$
ICube (UMR 7357 CNRS), Universit{\'e} de Strasbourg, France.
}%
\thanks{$^{3}$
IHU, Institut Hospitalo-Universitaire, Strasbourg, France.
}%
\thanks{$^{4}$
Instituto de Investigaci{\'o}n en Ingenier{\'i}a de Arag{\'o}n (I3A), Universidad de Zaragoza, Spain.
{\tt\small josemari@unizar.es}}%
}

\begin{document}

\maketitle
\thispagestyle{empty}
\pagestyle{empty}

\begin{abstract}

Recovering surgical scene structure in laparoscope surgery is crucial step for surgical guidance and augmented reality applications. In this paper, a quasi dense reconstruction algorithm of surgical scene is proposed. This is based on a state-of-the-art SLAM system, and is exploiting the initial exploration phase that is typically performed by the surgeon at the beginning of the surgery. We show how to convert the sparse SLAM map to a quasi dense scene reconstruction, using pairs of keyframe images and correlation-based featureless patch matching. We have validated the approach with a live porcine experiment using Computed Tomography as ground truth, yielding a Root Mean Squared Error of 4.9mm.
\end{abstract}

\vspace{-1mm}
\section{INTRODUCTION}
There is much ongoing research to develop Augmented Reality (AR) surgical guidance systems for improving laparoscopic surgery. The idea is to allow hidden structures such as organ vessels and tumors to be visualized in real-time, by registering imaging data from different modalities such as CT or MRI \cite{Puerto-Souza,Sylvain,Collins2017}. An important research objective is to register the second modality with the laparoscopic images automatically without the use of artificial tracking markers or external tracking equipment. Feature-based solutions were proposed \cite{Puerto-Souza,Haouchine}, however they fail in the case of textureless tissues where feature detection is extremely unreliable. Feature tracking \cite{Haouchine,KLT} is very fragile at textureless regions, and can suffer from drift and drop-off due to occlusion, sudden camera motion and motion blur. Furthermore feature-based solutions do not produce a dense 3D scene reconstruction, which is necessary for accurate inter-modal registration \cite{Collins2017}.

Several computer vision based solutions have been proposed for 3D reconstruction of laparoscopic scenes. Structure-from-Motion (SfM) \cite{Sun,Hu} is a well-established approach, however it requires offline batch processing and is not suitable for real-time application. Shape from Shading \cite{Collins2012} exploits the shading effect to deduce the scene's 3D structure, however it is weakly constrained problem and require multiple depth cues in a hybrid method, such as SfM/SfS as concluded by Collins et al. \cite{Collins2012}. Stereo-vision based solutions were also proposed \cite{Lin,Stoyanov2010} for dense scene reconstruction, however they are not applicable to monocular laparoscopes, which are by far the most common type.

Recently, SLAM (Simultaneous Localization And Mapping) systems have emerged as an excellent approach to reconstruct laparoscopic scenes and to compute the laparoscope's 3D pose in real-time. Typically, in the laparoscopic surgery an initial exploration is performed in order to explore the abdominal cavity. During exploration the surgeon does not manipulate the environment with tools, so the scene can be assumed to be rigid. This has been developed for both stereo  \cite{Mountney1,Mountney2} and monocular laparoscopes \cite{Grasa,Nader}. ORB-SLAM \cite{Raul} in particular shows remarkable tracking performance but at the expense of low map density \cite{Nader}. This is because it only reconstructs the 3D positions of sparse ORB features that have been matched in two or more keyframes.

We propose a SLAM based quasi dense reconstruction algorithm which is able to reconstruct the surgical environment using only a monocular endoscope and no extra tracking equipment. It works by densifying a sparse reconstructed map computed during exploration phase by e.g. ORB-SLAM, using pairs of keyframe images and correlation-based featureless patch matching. Only a small number of relevant keyframe pairs are selected, therefore keeping computation time low. Keyframe pairs are selected using their respective baseline in the covisibilty graph, and are treated as a stereo pair. Densification is then done in three main steps: \textit{initial feature based densification} where we do a 3D reconstruction of unmatched features (cf. Sec. \ref{initial_map_densification_section}), \textit{depth propagation} where we propagate the reconstruction to featureless regions (cf. Sec. \ref{depth_propagation_section}) and finally \textit{reconstruction post-processing}, where outliers are removed and the reconstruction is smoothed (cf. Sec. \ref{postprocessing_step}).

Semi-dense SLAM has been proposed in \cite{Engel,Mur-Artal} limiting the dense reconstruction to highly textured image areas. In contrast we densify regions with low contrast and without planarity assumption in low gradient regions as proposed in \cite{Alejo}. The advantage of densifying the sparse map after exploration phase is to maintain the per-frame computation time during SLAM, which is important to achieve reliable and robust tracking before and after the dense map  is estimated. Furthermore, unlike \cite{Alejo} we use densification based on Normalized Cross Correlation (NCC), which handles significant illumination changes that are common in surgical scenes \cite{Stoyanov2010}. The resulting quasi dense reconstruction can be used to accurately register pre-operative data to laparoscope's images, which is a vital component for AR surgical guidance.

\section{Quasi dense 3D scene reconstruction}

\subsection{Method overview}

\begin{figure}[thpb]
      \centering
      \includegraphics[scale=0.5]{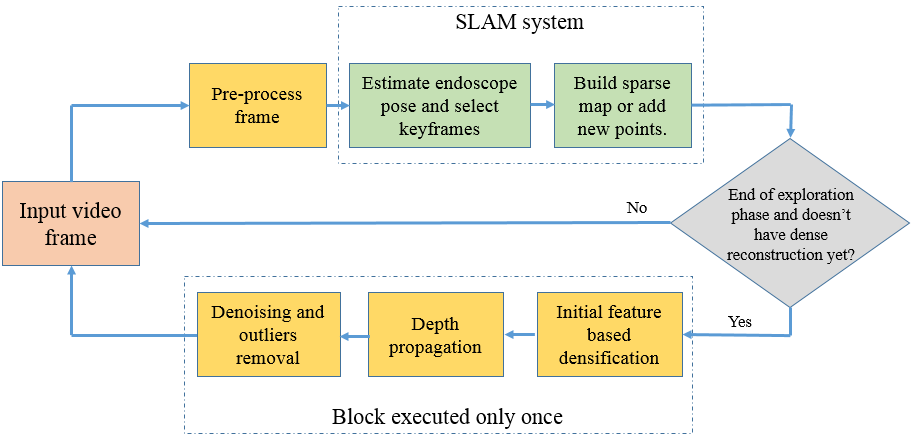}
      \caption{Algorithm pipeline.}
      \label{fig:reconstruction_pipeline}
   \end{figure}

We outline the proposed method in Figure \ref{fig:reconstruction_pipeline}.  We assume the laparoscope is pre-calibrated with fixed intrinsic using \cite{Zhang} immediately before exploration. During both SLAM and reconstruction densification we pre-process each frame to handle particular challenges of laparocopic image data (cf. Sec. \ref{pre-processing_step}). During exploration, SLAM is run until the end of this phase, which typically lasts no more than a minute. We use ORB-SLAM but any good feature-based SLAM approach could be used. The SLAM process is denoted by the top loop in Figure \ref{fig:reconstruction_pipeline}. This outputs a set of keyframes, their respective camera poses, a set of features detected in each keyframe and sparse 3D map. Next the three stages: feature based densification, depth propagation and reconstruction post-processing are run. Once finished, the laparoscope's pose can be tracked in the incoming laparoscopic frames in real-time using the sparse ORB-SLAM map.

\vspace{-2mm}


\vspace{-2mm}
\subsection{Frame pre-processing}
\label{pre-processing_step}
We first detect and eliminate specular reflections to avoid introducing false features to the SLAM system. This is done by converting the RGB frame to HSV and thresholding the saturation component. All detected features in these areas are ignored. Most feature detectors (including ORB) work on monochrome frames. We compute these by converting the RGB frame to monochrome using the average of the green and blue channels. This is because they give the highest contrast for human tissue \cite{Tromberg}.

\subsection{Building the keyframe neighborhood graph}
We construct a neighborhood graph $G$ that connects pairs of SLAM keyframes. This is a sparse graph with typically $O(g)$ edges, where $g$ is the number of keyframes. Sparseness is necessary to keep processing time low. Each edge $(i,j)$ in $G$ corresponds to a stereo pair, and we use this for both feature-based densification and depth propagation. This is constructed as follows. For each keyframe $i$ we compute the stereo baseline $\alpha$ with respect to all neighbor keyframes as a ratio between sparse map median depth and the distance between the two keyframes. We add an edge $(i,j)$ if $\alpha$ falls within the range  $\alpha_0 \leq \alpha \leq \alpha_1$. The lower-bound $\alpha_0$ ensures there is sufficient baseline with which to reliably reconstruct points in 3D. The upper-bound $\alpha_1$ ensures that the keyframes are not too far from each other. We give the default values of $\alpha_0$ and $\alpha_1$ (and all other parameters defaults) in Table \ref{table:1}.

\subsection{Feature based densification}
\label{initial_map_densification_section}
\begin{figure}[thpb]
      \centering
      \includegraphics[scale=0.5]{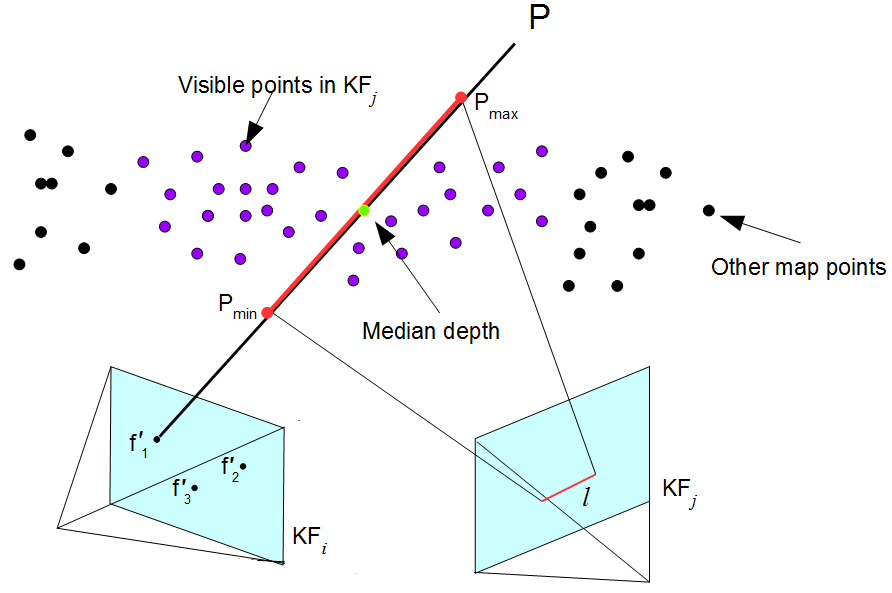}
      \caption{Epipolar guided search. ${KF_i}$ and ${KF_{j}}$ are current keyframe and its neighbor keyframe, respectively.}
      \label{epipolar_geometry_fig}
   \end{figure}

We process each keyframe pair $(i,j)\in G$ as follows. We have two types of features detected in $i$: matched {\textit{f}} and unmatched features {\textit{f'}}. The matched features are those that have been already matched by the SLAM system in another keyframes, have been triangulated, and have been inserted into the map. Our goal is to reconstruct each feature in {\textit{f'}}. The process works by a cross-correlation search guided by epipolar geometry as illustrated in Fig. \ref{epipolar_geometry_fig}. This works by searching exhaustively over an epipolar line segment $l$  with a margin of 10 pixels using NCC matching with a $N$ X $N$ window. We then triangulate the matches and filter them for removing outliers according to three criteria: Matches with NCC score less than a threshold $\tau$ are eliminated. Matches with negative depths are eliminated. Matches with a parallax angle lower than $\gamma$ are eliminated. The triangulated 3D position of each matched feature that passes the filters is then inserted into the map.

We bound the length of the epipolar segment $l$ to keep computation cost low using median depth of all visible points in keyframe $j$. Two extreme points on the back-projected ray are used to bound $l$ which are ${P_{min}}$ and ${P_{max}}$. The two points depths are computed by averaging and doubling median depth of visible map points, respectively.

\vspace{-5mm}
\subsection{Featureless depth propagation}
\label{depth_propagation_section}
After feature-based densification, we further densify the map at featureless regions through a depth propagation algorithm. The process works on each keyframe pair $(i,j)\in G$ as follows. First we take all points that were matched in keyframes $i$ and $j$, and use their depths as seed depths which are then propagated to neighboring pixels. We then continue to propagate depth around seeds on best-first basis by popping a seeds queue, as proposed in \cite{Stoyanov2010}. New matches are added to the queue as the algorithm iterates until no more matches can be popped.

Consider a seed point with a 2D position $m$ in keyframe $i$ and $m'$ in keyframe $j$, with $N(m)$ and $N(m')$ spatial neighbored pixels, respectively in a $6 \times 6$ window. These seed matches are used to control the smoothness of the disparity estimation of all $N(m)$ and $N(m')$ pixels. For each neighbored pixel $u \in N(m)$ a NCC is used to find a corresponding match $u'$ in a $6 \times 6$ window centered in the corresponding spatial location in keyframe $j$, that has higher NCC score than $\tau$ and satisfy the smoothness constrain defined in eq. (1). We use $\beta$ to control the smoothness of the disparity estimation. NCC is used as a similarity measure during propagation step.

\vspace{-4mm}
\begin{equation}
\begin{split}
N(m,m') = \left\lbrace (u,u'), ||(u-u')-(m-m')|| \leqslant \beta \right\rbrace 
\end{split}
\end{equation}


\subsection{Outlier removal and denoising}
\label{postprocessing_step}
Because depth propagation operates on keyframe pairs, there will be some disagreement due to noise across different keyframe pairs, typically at very low-textured regions. We deal with this by a robust averaging and merging. First we detect any remaining outliers in the quasi dense map using point neighborhood statistics \cite{Rusu}. This works by eliminating points if they are unusually far apart from the nearest $\kappa$ points, according to a threshold $\rho$ multiplied by standard deviation of all points distances. We then remove noise using Moving Least Square (MLS) \cite{Alexa}. This works by fitting a local plane to each point using all $\eta$ nearest neighbors. The denoised point is then computed by orthogonally projected onto the fitted plane.


\section{Experimental results}
\subsection{Experimental design}
\begin{figure}[H]
      \centering
      \subfigure[]{
          \includegraphics[width=0.20\textwidth]{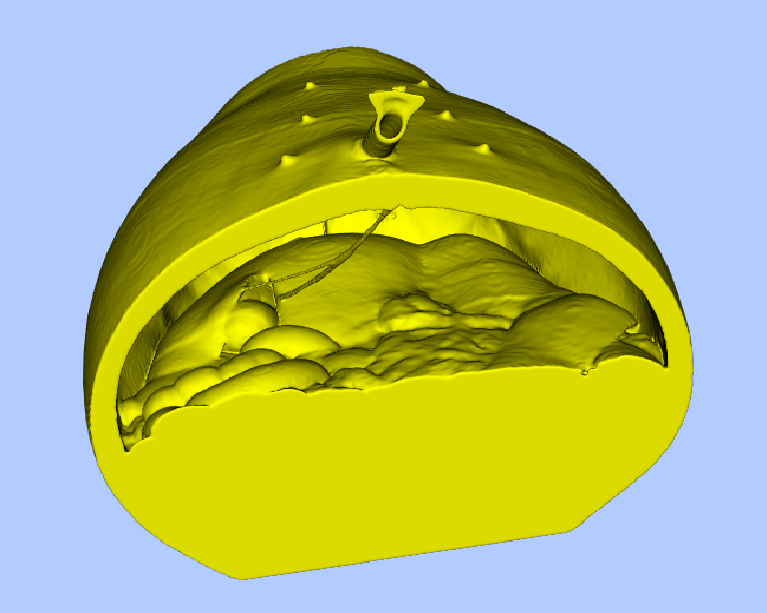}
      }
      \subfigure[]{
          \includegraphics[width=0.20\textwidth]{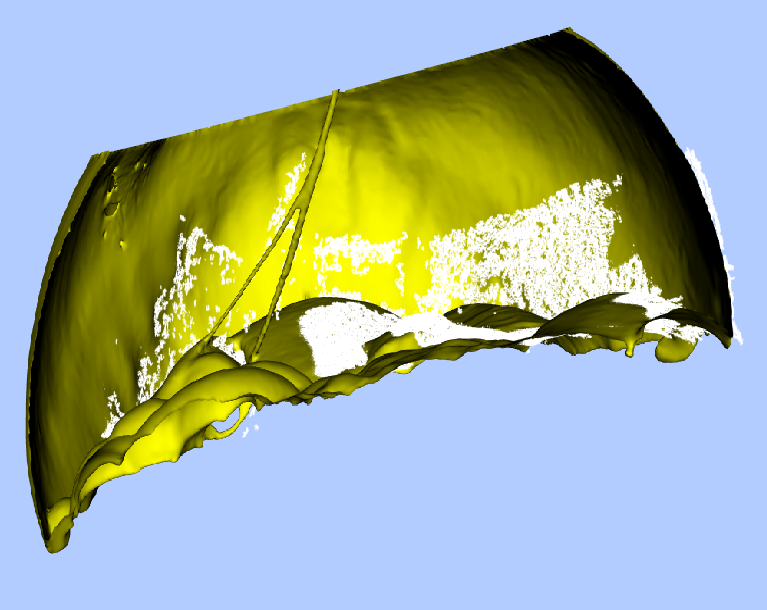}
      }\\
      \vspace{-2mm}
      \subfigure[]{
          \includegraphics[width=0.20\textwidth]{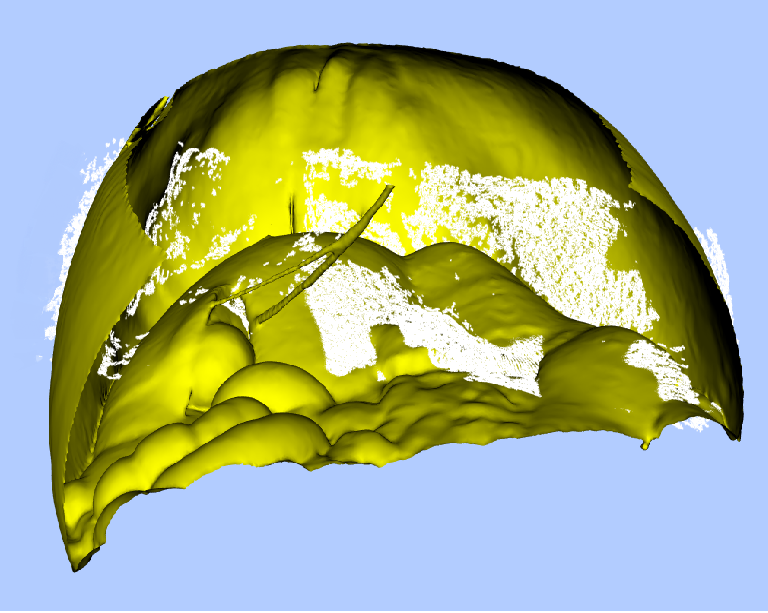}
      }     
      \subfigure[]{
          \includegraphics[width=0.20\textwidth]{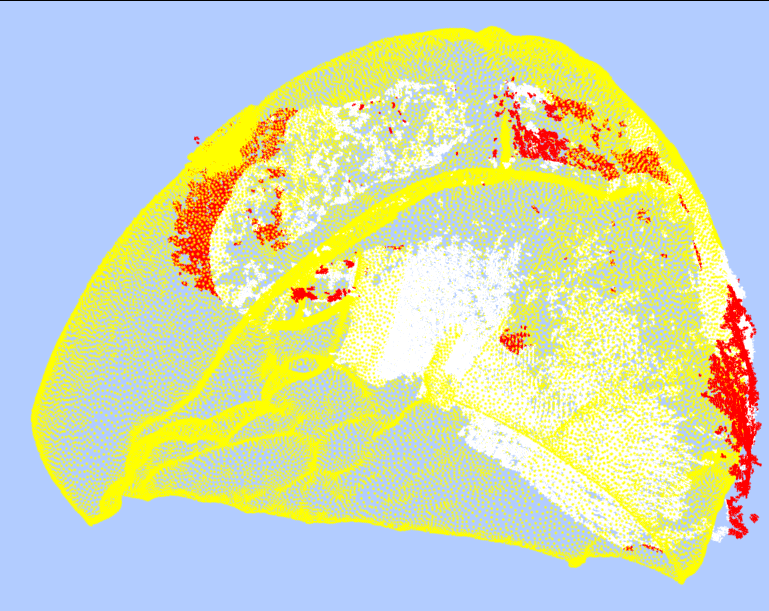}
      }
      \caption{Alignment of reconstructed dense map. (a) Whole CT surface of the pig abdominal cavity. (b,c) ICP alignment from two viewpoints with the visible part of CT surface in the laparoscope images. (d) yellow, white and red points are CT model points, inliers and outliers map points, respectively.}
      \label{fig:RMSE_fig}
   \end{figure}

\begin{figure*}[h]
      \centering
        \includegraphics[height=4.5cm]{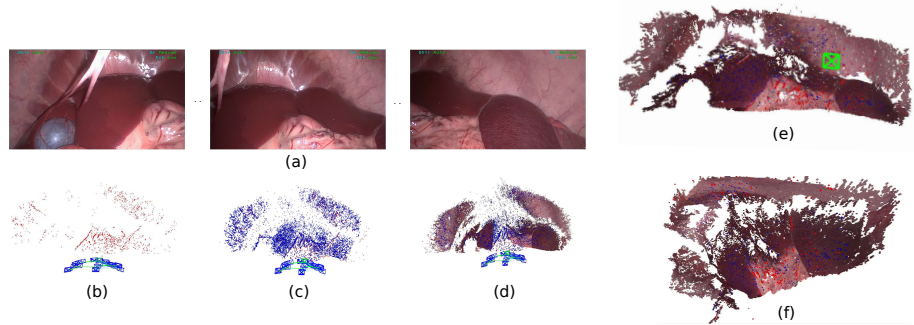}       
      \caption{Quasi dense reconstruction. (a) 3 Frames from the sequence. (b) normal ORB-SLAM map. (c) feature based densification. (d) depth propagation. (e,f) reconstructed scene from two different direction.}
      \label{fig:res1}
\end{figure*}

We evaluated our approach with a live in-vivo porcine experiment using a CT scan as ground-truth. In this experiment we reconstructed a porcine liver surrounding abdominal viscera, abdominal wall and diaphragm (cf. Fig. \ref{fig:res1}(e,f)) from 10 seconds exploratory video. Example frames are shown in Fig. \ref{fig:res1}(a). A CT was then acquired during 10 second expiration breath hold and is manually segmented by an expert to generate a 3D volume with 0.876mm x 0.876mm x 0.799mm voxel size and 749, 318 vertices, Fig. \ref{fig:RMSE_fig}(a).

Fig. \ref{fig:res1} shows the results of each step in reconstruction algorithm. Fig. \ref{fig:res1}(b) shows the sparse ORB-SLAM map from the initial exploration, where map points are shown in red and keyframes poses are the blue rectangles. Fig. \ref{fig:res1}(c), shows the newly added points, in blue after the feature based densification step. Fig. \ref{fig:res1}(d) shows the reconstruction after the featureless densification stage. Fig. \ref{fig:res1}(e,f) shows the reconstruction from different view points after outlier removal and denoising with normal ORB map point highlighted in red and the newly added points from feature based densification stage are highlighted in blue. Note that there are holes in the reconstruction due to specular reflections and regions of extremely homogeneous texture. More details can be appreciated at our video \cite{video1_myliver}. Another video \cite{video2_HamlynLiver} shows the reconstruction results of one liver sequence available at Hamlyn Centre Laparoscopic/Endoscopic Video Datasets \cite{HamlynDataset}.


\vspace{-2mm}
\begin{table}[h!]
\centering
\caption{Tuning parameters}
\begin{tabular}{||c | c | c | c | c | c | c| c | c ||}
 \hline
 $\alpha$ & $N$ & $\tau$ & $\gamma$ & $\beta$ & $\rho$ & $\kappa$ & $\eta$ \\ [0.5ex]
 \hline\hline
 [0.05,0.09] & 19 & 0.3 & [$2.14^{\circ}$,$90^{\circ}$] & 1.5 & 4 & 40 & 40 \\ [1ex]
 \hline
\end{tabular}

\label{table:1}
\end{table}

\vspace{-3mm}
\subsection{Implementation details and computation time}
The system is implemented in C++ with OpenCV and PCL libraries and executed on desktop PC with Intel Core i7 CPU @ 2.6 GHz and 4GB RAM. The average tracking time is 23ms with 1600x900 image resolution. Average time required for: feature based densification was 5ms per keyframe, depth propagation was 25ms per keyframe (time for matching and triangulating points), MLS denoising was 1min due to computing normal for each points and polynomial fitting. The total number of reconstructed points was 348, 068 point. After obtaining dense scene reconstruction, the endoscope pose is then tracked in 25ms on average because only the sparse ORB-SLAM map is considered.
\begin{figure}[H]
      \centering
      \subfigure[]{
          \includegraphics[width=0.22\textwidth]{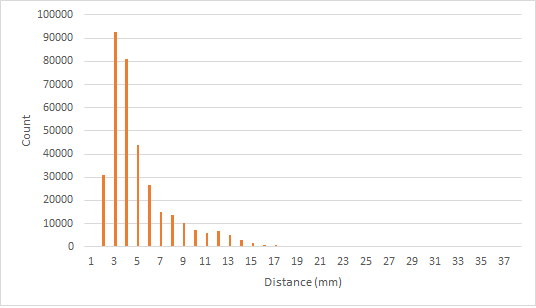}
	      }
      \subfigure[]{
          \includegraphics[width=0.22\textwidth]{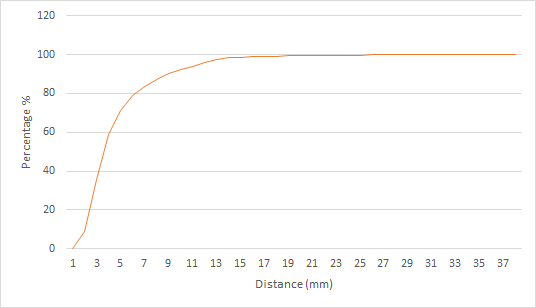}
      }
      \caption{Distances Distributions.}
      \label{fig:Histograms_fig}
   \end{figure}
\vspace{-6mm}

\vspace{-5mm}
\subsection{Registering the map to CT and measuring accuracy}
As with any SLAM system, our reconstruction is up to a similarity transform (i.e. an arbitrary scale and rigid coordinate transform). To evaluate accuracy we aligned the reconstruction to the CT model (cf. Fig. \ref{fig:RMSE_fig}(a)) using a best-fitting similarity transform. This was found by manually selecting 3 landmarks to roughly estimate the scale and initial orientation by Horn's algorithm \cite{Horn}. Then Iterative Closest Point (ICP) was run until convergence to refine the initial alignment. Fig. \ref{fig:RMSE_fig} (b,c) shows the alignment between CT model in yellow and reconstructed dense map in white.

Accuracy was measured by the euclidean distance of each map point to its closest point on the CT model's surface. The Root Mean Squared Error (RMSE) was used to evaluate the overall error, which was 4.9mm. Fig. \ref{fig:Histograms_fig}(a,b) shows distances distributions of total number of map points besides the accumulative histogram. It can be seen that 85\% of points with lower distances than 6.7mm. Thus, thresholding errors lower than 6.7mm as inliers (85\%) and the rest as outliers(15\%), (cf. Fig. \ref{fig:RMSE_fig}(d)) reduces the RMSE error to 2.8mm. Those 15\% outliers were abdominal wall points and it is quite hard to correctly reconstruct them due to non-rigid deformation by the breathing cycle which prohibits correct re-projection \cite{Marcinczak}.

\vspace{-1mm}
\section{CONCLUSIONS}
We have presented a simple approach for quasi-dense 3D reconstruction of laparoscopic scenes, that robustly densifies a sparse SLAM reconstruction. Because densification is embedded in SLAM, we keep all the benefits of state-of-the-art feature-based SLAM system such a ORB-SLAM, including fast tracking, mapping and automatic relocalization. Our preliminary results on an in-vivo porcine dataset are very promising, with a RMSE of 4.9mm. Future directions include improving reconstruction accuracy by including multiple views in the reconstruction process of the same point. Reducing dense reconstruction computation time, especially the denoising stage. Additionally, we aim to use the reconstruction for automatic registration with a CT model. 
\addtolength{\textheight}{-12cm}   



\vspace{-1mm}
\section*{ACKNOWLEDGMENT}
This work is part of a project of the Investissements d'Avenir program ("Investing in the Future") called 3D-Surg, funded by BPIfrance. It is also partially funded by the Spanish government DPI2015-67275-P and Aragonese DGA T04-FSE.


\vspace{-3mm}
\begin{thebibliography}{}

\bibitem{Puerto-Souza}
G. Puerto-Souza, J. A. Cadeddu, and G. Mariottini, Toward long-term and accurate augmented-reality for monocular endoscopic videos. IEEE Trans. Bio. Eng., vol. 61(10), pp. 2609-20, 2014.
\bibitem{Sylvain}
S. Bernhardt, S. A. Nicolau, L. Soler, C. Doignon, The status of augmented reality in laparoscopic surgery as of 2016. Medical Image Analysis, vol.37, pp. 66-90, 2017.
\bibitem{Collins2017}
T. Collins, P. Chauvet, C. Debize, D. Pizarro, A. Bartoli, M. Canis, N. Bourdel,  A System for Augmented Reality Guided Laparoscopic Tumour Resection with Quantitative Ex-vivo User Evaluation, CARE 2016.
\bibitem{Haouchine}
N. Haouchine, J. Dequidt, M. Berger, and S. Cotin, Monocular 3d reconstruction
and augmentation of elastic surfaces with self-occlusion handling. IEEE Trans.
Vis. Comput. Graph, vol 21(12), pp. 1363-1376, 2015.
\bibitem{KLT}
C. Tomasi and T. Kanade, Detection and Tracking of Point Features. Carnegie Mellon University Technical Report CMU-CS-91-132, 1991.
\bibitem{Sun}
D. Sun, J. Liu, C.-A. Linte, H. Duan, R.-A. Robb, Surface Reconstruction from Tracked Endoscopic Video Using the Structure from Motion Approach. MIAR 2013, pp. 127-135, 2013
\bibitem{Hu}
M. Hu, G. Penney, M. Figl, P. Edwards, F. Bello, R. Casula, D. Rueckert, D. Hawkes, Reconstruction of a 3D surface from video that is robust to missing data and outliers: application to minimally invasive surgery using stereo and mono endoscopes. Med. Image Anal, vol. 16(3), pp. 597-611, 2012.
\bibitem{Collins2012}
T. Collins, A. Bartoli, Towards Live Monocular 3D Laparoscopy using Shading and Specularity. IPCAI 2012, pp 11-21, 2012
\bibitem{Lin} 
B. Lin, A. Johnson, X. Qian, J. Sanchez, Y. Sun, Simultaneous  Tracking,  3D Reconstruction and  Deforming  Point  Detection  for  Stereoscope  Guided  Surgery. MICCAI, pp. 35-44, 2013.
\bibitem{Stoyanov2010}
D. Stoyanov, M.V. Scarzanella, P. Pratt,  G.Z. Yang, Real-time stereo reconstruction in robotically assisted minimally invasive surgery, MICCAI, pp. 275–282, 2010
\bibitem{Mountney1}
P. Mountney, D. Stoyanov, A.J. Davison, G.Z. Yang, Simultaneous stereoscope localization  and  soft-tissue  mapping  for  minimal  invasive  surgery. MICCAI 2006, Part I. LNCS, vol. 4190, pp. 347-354, 2006
\bibitem{Mountney2}
P. Mountney, G.-Z. Yang, Motion compensated SLAM for image guided surgery. MICCAI, pp. 496-504, 2010
\bibitem{Grasa}
O. G. Grasa, E. Bernal, S. Casado, I. Gil, J.M.M Montiel, Visual  SLAM  for handheld monocular endoscope. IEEE Trans. on Med. Imag., vol. 33(1), pp. 135-146, 2014.
\bibitem{Nader} 
N. Mahmoud, I. Cirauqui, A. Hostettler, C. Doignon, L. Soler, J Marescaux, J. M. M. Montiel, ORBSLAM-based Endoscope Tracking and 3D Reconstruction, CARE, pp. 72-83, 2017
\bibitem{Raul}
R. Mur-Artal, J. M. M. Montiel, J. D. Tardos, ORB-SLAM: A Versatile and Ac-
curate Monocular SLAM System. IEEE Trans. on Robotics, vol. 31(5), pp. 1147-1163, 2015
\bibitem{Engel}
J. Engel, T. Schops, and D. Cremers, LSD-SLAM: Large-scale direct monocular SLAM.  ECCV, pp. 834-849, 2014
\bibitem{Mur-Artal}
R. Mur-Artal and J.D. Tard\'os, Probabilistic Semi-Dense Mapping from Highly Accurate Feature-Based Monocular SLAM. RSS 2015.
\bibitem{Alejo}
A. Concha and J. Civera, DPPTAM: Dense piecewise planar tracking and mapping from a monocular sequence. IROS 2015, pp. 5686-5693. 2015.
\bibitem{Zhang}
Z. Zhang. A Flexible New Technique for Camera Calibration. IEEE Trans. Pattern
Anal, vol. 22(11), pp. 1330-1334, 2000.
\bibitem{Tromberg}
B. J. Tromberg, N. Shah, R. Lanning, A. Cerussi, J. Espinoza, T. Pham, L. Svaasand, J. Butler, Non-Invasive In Vivo Characterization of Breast Tumors Using Photon Migration Spectroscopy. Neoplasia, vol. 2(1-2), pp.26-40, 2000
\bibitem{Rusu}
R. B. Rusu, Z. C. Marton, N. Blodow, M. Dolha, and M. Beetz, Towards 3D Point cloud based object maps for household environments. Robotics and Autonomous Systems, vol. 56(11), pp. 927–941, 2008
\bibitem{Alexa}
M. Alexa, J. Behr, D. Cohen-Or, S. Fleishman, D. Levin, C. T. Silva, Computing and Rendering Point Set Surfaces. IEEE Trans. on Vis. and Comp. Graph., vol. 9(1), 2008
\bibitem{video1_myliver}
Youtube. (2017, May 25). SLAM based Quasi Dense Reconstruction For Minimally Invasive Surgery Scenes (Private Dataset)) [Video file]. Retrieved from https://www.youtube.com/watch?v=HQmVRSNFVu0\&feature=youtu.be
\bibitem{video2_HamlynLiver}
Youtube. (2017, May 25). SLAM based Quasi Dense Reconstruction For Minimally Invasive Surgery Scenes (Hamlyn Dataset) [Video file]. Retrieved from https://www.youtube.com/watch?v=oG54CBzqVh0\&feature=youtu.be
\bibitem{HamlynDataset}
London, I.C., Hamlyn centre laparoscopic / endoscopic video datasets (2017). URL http://hamlyn.doc.ic.ac.uk/vision/. [Accessed 24 May 2017]
\bibitem{Horn}
B.K. Horn, Closed-form solution of absolute orientation using unit quaternions. J OPT SOC AM A, vol. 4(4), pp. 629-642, 1987
\bibitem{Marcinczak}
J.M. Marcinczak, R.R. Grigat, Total Variation Based 3D Reconstruction from Monocular Laparoscopic Sequences.  ABD-MICCAI, pp 239-247, 2014.


\end{thebibliography}
\end{document}